\documentclass[onefignum,onetabnum]{siamonline190516}

\usepackage{amsfonts, amsmath, amssymb}
\usepackage{graphbox,graphicx}
\usepackage{epstopdf}
\usepackage[algo2e,linesnumbered,boxed]{algorithm2e}
\usepackage{dsfont}
\usepackage{booktabs}
\usepackage{todonotes}
\usepackage{mathtools}
\usepackage{tikz}
\usepackage{wrapfig}

\ifpdf
  \DeclareGraphicsExtensions{.eps,.pdf,.png,.jpg}
\else
  \DeclareGraphicsExtensions{.eps}
\fi

\usepackage{enumitem}
\setlist[enumerate]{leftmargin=.5in}
\setlist[itemize]{leftmargin=.5in}

\newsiamremark{remark}{Remark}
\newsiamremark{hypothesis}{Hypothesis}
\crefname{hypothesis}{Hypothesis}{Hypotheses}
\newsiamthm{claim}{Claim}

\headers{Robust Risk-Aware RL}{Jaimungal, Pesenti, Wang, Tatsat}

\title{Robust Risk-Aware Reinforcement Learning \thanks{Opinions expressed within the content are solely the authors' and do not necessarily reflect the opinions and beliefs of their affiliates. The authors thank the anonymous referees for useful comments that ultimately improved the presentation of the paper.
\funding{SJ and SP would like acknowledge support from the Natural Sciences and Engineering Research Council of Canada (grants RGPIN-2018-05705, RGPAS-2018-522715, and DGECR-2020-00333, RGPIN-2020-04289).}}}

\author{
Sebastian Jaimungal
\thanks{Dept. Statistical Sciences, University of Toronto
  (\email{sebastian.jaimungal@utoronto.ca}, \url{http://sebastian.statistics.utoronto.ca}, \email{silvana.pesenti@utoronto.ca},
  \url{https://utstat.toronto.edu/pesenti/}, \email{yesheng.wang@mail.utoronto.ca}).}
\and 
Silvana M. Pesenti
\footnotemark[2]
\and 
Ye Sheng Wang\footnotemark[2]
\and 
Hariom Tatsat\thanks{Barclays Capital (\email{hariom.x.tatsat@barclays.com}).}
}

\usepackage{amsopn}

\tikzstyle{reward}=[shape=circle,draw=blue!50,fill=blue!10]
\tikzstyle{action}=[shape=circle,draw=green,fill=green!10]
\tikzstyle{state}=[shape=circle,draw=red!50,fill=red!10]
\tikzstyle{gru}=[shape=rectangle,draw=black!50,fill=lime!10]
\tikzstyle{obs}=[shape=circle,draw=blue!50,fill=blue!10]
\tikzstyle{lightedge}=[<-,dotted]
\tikzstyle{mainstate}=[state,thick]
\tikzstyle{mainedge}=[<-,thick]

\newcommand{\RM}{{\mathcal{R}}}
\newcommand{\U}{{\mathcal{U}}}
\newcommand{\A}{{\mathcal{A}}}
\newcommand{\F}{{\mathcal{F}}}
\newcommand{\D}{{\mathcal{D}}}

\renewcommand{\Finv}[1]{{F^{-1}_{#1}}}

\newcommand{\R}{{\mathds{R}}}

\newcommand{\Lp}{{\mathbb{L}^p}}

\newcommand{\E}{{\mathbb{E}}}
\renewcommand{\P}{{\mathbb{P}}}

\newcommand{\ep}{\varepsilon}

\newcommand{\Id}{{\mathds{1}}}

\newcommand{\mfp}{{\mathfrak{p}}}

\newcounter{example}[section]
\newenvironment{example}[1][]{\refstepcounter{example}\par\medskip
   \noindent \textbf{Example~\theexample. #1} \rmfamily
   \begin{small}}{\end{small}\medskip}
   
 \DeclareMathOperator{\sgn}{sgn}

\newcommand{\new}[1]{#1}

\usepackage{wrapfig}

\graphicspath{{./Figures/}}

\allowdisplaybreaks

\begin{document}

\maketitle

\begin{abstract}
We present a reinforcement learning (RL) approach for robust optimisation of risk-aware performance criteria. To allow agents to express a wide variety of risk-reward profiles, we assess the value of a policy using rank dependent expected utility (RDEU). RDEU allows the agent to seek gains, while simultaneously protecting themselves against downside risk. To robustify optimal policies against model uncertainty, we assess a policy not by its distribution, but rather, by the worst possible distribution that lies within a Wasserstein ball around it. Thus, our problem formulation may be viewed as an actor/agent choosing a policy (the outer problem), and the adversary then acting to worsen the performance of that strategy (the inner problem). We develop explicit policy gradient formulae for the inner and outer problems, and show its efficacy on three prototypical financial problems: robust portfolio allocation, optimising a benchmark, and statistical arbitrage.
\end{abstract}



\section{Introduction}
Many problems in financial mathematics, economics, and engineering may be cast in the form of a stochastic optimisation problem, and typically the agent's optimal control depends on the underlying (dynamic) model assumptions. Models, however, are approximations of the world, whether they are purely data driven (i.e., empirical) models, parametric models estimated from data, or models that are posited to reflect the given stochastic dynamics. As models are approximations, understanding how to protect ones decisions from  uncertainty inherent in a model is of paramount importance. Thus, here we consider robust stochastic optimisation problems where the agent chooses the action that is optimal under the worst case within an uncertainty set.

In many contexts, and particularly so in financial modelling, it is important to account for risk. Using expected utility of rewards is one approach for trading off risk and reward, however, there are many models of decision under uncertainty that go beyond it \cite{Yaari1987Economitrica}, \cite{pflug2007ambiguity},\cite{Bernard2020robust}. Here, we take the \new{rank} dependent expected utility (RDEU) framework of \cite{Yaari1987Economitrica} which allows agents to account not only for the concavity in their utility, but also allows them to distort the probabilities of outcomes to better accommodate empirical violations of expected utility theory \cite{diecidue2001intuition}.

While only specific examples of robust stochastic optimisation problems admit (semi-) analytical solutions or are numerically tractable, a general framework for solving robust stochastic optimisation problems is still missing and that is the focus of this paper. Hence, we develop a reinforcement learning (RL) approach for solving a general class of robust stochastic optimisation problems, where agents aim to minimise their risk --  measured by RDEU -- subject to model uncertainty, and thus robustifying their actions.

In our setting, an agent's action induces a univariate controlled random variable (rv) which is subject to distributional uncertainty, modelled via the Wasserstein distance. Notable is that in our setting, while the uncertainty is on the controlled rv and the alternative distributions lie within a Wasserstein ball around it, the alternate distributions may also have other structural constraints.

The vast majority of the literature on RL considers maximising expected total reward, while in many contexts, and particularly so in financial modelling, it is important to account for risk. Any RL approach that accounts for risk is termed risk-aware or risk-sensitive RL. While risk-aware RL often focuses on expected utility \cite{mihatsch2002risk,fei2020risk},  or probability of entering undesirable states \cite{geibel2005risk}, recently \cite{tamar2015policy} extended policy gradient methods to account for coherent measures of risk (see \cite{huang2021convergence} for proof of convergence). Here, however, we are interested in RDEU measures of risk that falls outside the class of coherent risk measures.

Distributionally robust RL looks at robustifying strategies against uncertainty in  the distribution of either transition probabilities or the random variable whose expectation one seeks to maximise.  A (risk-neutral) distributionally robust RL approach for Markov decision processes, where robustness is induced by looking at all transition probabilities (from a given state) that have relative entropy with respect to (wrt) a reference probability less than a given epsilon, is developed in \cite{smirnova2019distributionally}.  \cite{abdullah2019wasserstein} develops a (risk-neutral) robust RL  paradigm where policies are randomised with a distribution that depends on the current state, see \cite{wang2020reinforcement} for a continuous time version of randomised policies  with entropy regularisation and \cite{guo2020entropy,firoozi2020exploratory} for its generalisation to mean-field game settings. In \cite{abdullah2019wasserstein}, the uncertainty is placed on the conditional transition probability from old state and action to new state, and the set of distributions are those that lie within an ``average'' $2$-Wasserstein ball around a benchmark model's distribution. As randomised policies are used, the constraint and policy decouple. In this work, there is no such decoupling and, moreover, we develop approaches for both deterministic and randomised policies.

As far as the authors are aware, this paper fills two gaps in the literature. The first is the incorporation of RDEU measures of risk to RL problems, and the second is  robustifying risk-aware RL. We fill these gaps by posing a generic robust risk-aware optimisation problem, develop policy gradient formulae for numerically solving it, and illustrate its tractability on three prototypical examples in financial mathematics.

The remainder of this paper is structured as follows: Section \ref{sec:optimisation} introduces the robust stochastic optimisation problem. Section \ref{sec:gradients}  provides the RL policy gradient formulae for the inner and outer problems. Section \ref{sec:example} illustrates the tractability of the RL framework on three examples: robust portfolio allocation, optimising a benchmark, and statistical arbitrage.

\section{Robust Optimisation Problems}\label{sec:optimisation}

We consider agents who measure the risk/reward of a rv using Yaari's dual theory \cite{Yaari1987Economitrica}. As Yaari argues, agents not only value outcomes according to an utility function, but also view probabilities of outcomes subjectively and thus distort them. This leads  to the notion of rank dependent expected utility (RDEU) defined below. 
\begin{definition}[RDEU]
The RDEU of a rv $Y$ may be defined via a Choquet integral as
\begin{equation}
\label{eqn: RDEU}
    \RM^U_g[Y] := 
  \int_{-\infty}^0 1 - g\big(\P (U(Y) > y)\big) \, dy 
  - \int_0^{+\infty}  g\big(\P (U(Y) > y)\big)\, dy \,,
\end{equation}
where $g \colon [0,1] \to [0,1]$ is an increasing function with $g(0) = 0$ and $g(1) = 1$, called distortion function, and $U$ is a non-decreasing concave utility function. We assume that $U$ is differentiable almost everywhere.
\end{definition}
The above definition assumes that positive outcomes correspond to gains and negative ones to losses. The RDEU framework subsumes the class of distortion risk measures, for $U(x) = x$, which includes the well-known Conditional-Value-at-Risk (CVaR), see Section \ref{sec:example}. Moreover, it includes the expected utility framework when $g(x) = x$, in which case $\RM^U_g[Y] = -\E[U(Y)]$. Throughout, we refer to the RDEU of a rv as the rv's risk.

We consider the situation where an agent's action $\phi \in \varphi$ induces a rv $X^\phi$ and that the agent aims to minimise the risk associated with $X^\phi$, i.e. $\RM^U_g[X^\phi]$. However, due, to the presence of model uncertainty -- distributional uncertainty on $X^\phi$ -- the agent, instead of choosing actions with minimal risk, chooses the action that minimises the worst-case risk of all alternative rvs $X^\theta$, where $\theta$ belongs to an uncertainty set $\vartheta_\phi$, which may depend on the agents' action $\phi$. Specifically, the agent aims to solve the robust optimisation problem
\begin{equation}
\inf_{\phi\in\varphi}\; \sup_{\theta\in\vartheta_\phi}\;\RM^U_{g}[X^\theta]
\;,
\qquad \text{where} \qquad
\vartheta_\phi:=\left\{\theta \in \vartheta : d_p[X^\theta,X^\phi]\le \ep\right\}\;,
    \tag{P}
    \label{eqn:P}
\end{equation}
where the admissible set of controls $\varphi\subseteq\R^m$, $X^\phi$ is a controlled $\R$-valued rv,  $X^\theta$ is an $\R$-valued rv parametrised by $\theta$, $\vartheta\subseteq R^n$ parametrises the robustness set, and $d_p[X,Y]$ denotes the $p$-Wasserstein distance between two rvs $X$ and $Y$, defined below. Problem \eqref{eqn:P} is only fully specified once the mappings $\phi\mapsto X^\phi$ and $\theta\mapsto X^\theta$ are given. The proposed RL approach allows for flexibility in these mappings, thus we make here, apart from the existence of a solution to  \eqref{eqn:P}, no further assumption on them. $X^\phi$ (and $X^\theta$) may result from sequences of decisions (multi-period) or a single decision (single period). We may  interpret \eqref{eqn:P} as an adversarial attack, where the agent picks an action, and an adversary distorts $X^\phi$ to have as worst performance as possible within a given Wasserstein ball. Below we provide several examples of problem \eqref{eqn:P} which we revisit in Section \ref{sec:example}.

Recall that the Wasserstein distance of order $p  \in [1, +\infty)$ between $X \stackrel{\P}{\sim} F_X$ and $Y \stackrel{\P}{\sim} F_Y$, is given by
(see e.g., \cite{ambrosio2003lecture}, Chap. 1)
\begin{equation}
d_p[X\,,\, Y] 
    := \inf_{\chi\in\Pi(F_{X},F_{Y})} \left(\int_{\R^2}|x-y|^p\,\chi(dx,dy)\right)^{\frac1p},
\end{equation}%
where $\Pi(F_{X},F_{Y})$ is the set of all bivariate probability measures  with marginals $F_{X}$ and $F_{Y}$. The $p$-Wasserstein distance defines a metric on the space of probability measures.

The robust stochastic optimisation problem \eqref{eqn:P} is a generalisation of distributional robust optimisation, where the uncertainty set is a subset of the space of distribution functions only, see e.g., \cite{esfahani2018data, Bernard2020robust}. Here, however,  the uncertainty set  $\vartheta_\phi$ possesses additional features in that it (a) may depend on the agent's action $\phi$, (b) the rv $X^\theta$ may have a structure induced by $\theta$, in which case not all rv within a Wasserstein distance around $X^\phi$ are feasible rvs, and (c) the set of feasible parameters $\theta$ belong to a set $ \vartheta$, which may impose additional constraints on $X^\theta$.

Problem \eqref{eqn:P}  performs a robust optimisation (over $\phi$) of $X^\phi$ as follows. Given $X^\phi$ from the ``outer'' problem, the ``inner'' problem $\sup_{\theta\in\vartheta_\phi} \RM_g^U[X^\theta]$ corresponds to a robust version of $X^\phi$'s risk. As $\ep\downarrow0$, the inner problem reduces to the RDEU of $X^\phi$. When $\ep>0$, however, the agent incorporates model uncertainty, and instead assesses the risk associated with $X^\phi$ by seeking over all alternate rv, generated by $\theta\in\vartheta$, that lie within a Wasserstein ball around it. 

\begin{example}[Robust Portfolio Allocation.]
\label{ex:Robust-Portfolio-Allocation} Suppose that $\varphi$ is the probability simplex in $d$-dimensions, for $\phi\in\varphi$ we write $X^\phi=\phi^\intercal X$, and $X=(X_1,\dots,X_d)$ represents the returns of $d$ traded assets. Further, let $\vartheta=\R^n$ and write $X^\theta=H_\theta(X^\phi)$, where $H_\theta(\cdot)$ is an artificial neural net (ANN) parameterised by $\theta$. In this setup, the inner problem corresponds to seeking over all distribution functions, that may be generated by the ANN and that lie within a Wasserstein ball around $X^\phi$; thus, the inner problem results in a robust estimate of the risk of $X^\phi$. The outer problem then seeks to find the best investment that is robust to model uncertainty. \cite{pflug2007ambiguity,esfahani2018data} investigate a similar class of problems, however, the uncertainty ball is on the inputs $X$ and not the output $X^\phi$ and they use coherent/convex risk measures as measures of risk compared to RDEU.
\end{example}

\begin{example}[Optimising Risk-Measures with a Benchmark.] 
\label{ex:benchmark}
Suppose that $\varphi$ is a singleton, the components of $\phi\in\varphi$ denote the percentage of wealth to invest in various assets, and $X^\phi$ denotes the terminal value of such an investment. Then, $X^\phi$ may be interpreted as benchmark strategy that the investor wishes to outperform in terms of RDEU. 

Let  $\theta\in\vartheta$ parameterise a dynamic self-financing trading strategy (e.g., parameters in an ANN that map time and asset prices to trading positions) whose terminal value is $X^\theta$. If we replace  in the inner problem in \eqref{eqn:P} the $\sup$ with $\inf$, the corresponding problem is to find a dynamic strategy that has the best risk of all portfolios within a Wasserstein ball around the benchmark. This example generalises \cite{pesenti2020portfolio} to the case of RDEU and also applies to incomplete markets.
\end{example}

\begin{example}[Robust Dynamic Trading Strategy.]
\label{ex:dynamic}
Consider the case where $X=(X_0,X_1,\dots,X_{T-1})$ denotes the price path of an asset at (trading) time points $0<t_1<t_2<\dots<t_{T-1}$, and $\varphi=[-a,a]^T$ denotes the shares bought/sold  at the sequence of trading times. For any $\phi\in\varphi$, the terminal wealth from the sequence of trades is
\begin{align}
    X^\phi 
    = -\sum_{i=0}^{T-1} \phi_i \;X_i + q_T^\phi\,X_T
    = \sum_{i=1}^T q_{i}^\phi\,(X_i-X_{i-1}),
\end{align}
where $q^\phi_i=\sum_{j=0}^{i-1} \phi_j$ is the total assets held at time $t_i$. We allow for the price to be affected by the trader's actions as described in Subsection \ref{sec:RobustStatArb_example}. Further, we set $X^\theta=H_\theta(X^\phi)$, where $H_\theta(\cdot)$ is an ANN parametrised by $\theta$. As in Example \ref{ex:Robust-Portfolio-Allocation} this corresponds to an agent who aims to minimise over $\phi\in\varphi$ a robust measure of risk of $X^\phi$. A related work is \cite{cartea2017algorithmic} who consider robust algorithmic trading problems using relative entropy penalisations under linear utility.
\end{example}

\section{Policy Gradients}\label{sec:gradients}

Policy gradient methods provide a sequence of policies/actions that improve upon one another by taking steps in the direction of the value function's gradient, where the gradient is taken wrt the parameters of the policy. In this section, we derive policy gradient update rules for optimising \eqref{eqn:P} over both $\phi$ and $\theta$. In Section \ref{sub-sec-random-action}, we provide a policy gradient formula when the agent controls not the action itself, but rather its distribution. Such actions are also referred to as relaxed controls, see \cite{wang2020reinforcement,firoozi2020exploratory,guo2020entropy}.

\subsection{The Inner Problem}

First, we study the inner problem of \eqref{eqn:P}. To do so, we employ an  augmented Lagrangian approach to incorporate the constraints. For this, we fix the rv $X^\phi$ and denote by $G_\phi$ and $G^{-1}_\phi$ its corresponding distribution, respectively, quantile function. We further denote the distribution and quantile function of $X^\theta$ by $F_\theta$ and $F^{-1}_\theta$, respectively. The augmented Lagrangian may then be written as
\begin{equation}
    L[\theta,\phi] = \RM_g^U[X^\theta]
    + \lambda\, c[X^\theta,X^\phi]
    + \tfrac{\mu}{2} (c[X^\theta,X^\phi])^2,
\end{equation}
where $c[X^\theta,X^\phi]:=\left(\left(d_p[X^\theta,X^\phi]\right)^p - \ep^p\right)_+$ is the $p$-Wasserstein constraint error, $(x)_+$ denotes the positive part of $x$, $\lambda$ is the Lagrange multiplier that enforces this constraint, and $\mu$ the penalty constraint.
The augmented Lagrangian approach fixes $\lambda$ and $\mu$, minimises/maximises $L[\theta,\phi]$, e.g., by using stochastic gradient descent (SGD), then updates $\lambda \leftarrow\lambda + \mu\,c[X^{\theta^*},X^\phi]$ and $\mu\leftarrow a\,\mu$ with some $a>1$. For an overview of the augmented Lagrangian approach see, e.g., \cite{birgin2014practical}, Chap. 4.

While the augmented Lagrangian may be estimated from a mini-batch of simulations, optimising over the parameters $\theta$ requires gradients wrt $\theta$. Many widely used risk measures, such as CVaR, RVaR, UTE, however, admit a derivative (of $g$) that has discontinuities, and whenever the derivative of $g$ has discontinuities, na\"ive back-propagation will incorrectly estimate its gradient. To overcome these potential discontinuities, we derive a gradient formula that can be estimated using mini-batch samples. 

\begin{proposition}[Inner Gradient Formula.]
\label{prop:inner-gradient-formula}
Let $X_c^\phi$ denote the version of $X^\phi$ that makes $(X^\theta,X_c^\phi)$ comonotonic -- i.e., reorder the realisations of $X_c^\phi$ according to the rank of $X^\theta$. If $g$ is left-differentiable, then
\begin{align}
\begin{split}
    \nabla_\theta L[\theta,\phi] =& \E\left[\left\{ U'\left(X^\theta\right)\gamma\left(F_\theta(X^\theta)\right)
    - p\,\Lambda\,
     |X^\theta - X_c^\phi|^{p-1} \sgn(X^\theta - X_c^\phi)\right\}
     \frac{\nabla_\theta F_\theta(x)|_{x=X^\theta}}{f_\theta(X^\theta)}
     \right]
     \label{eqn:L-gradient-inner}
\end{split}
\end{align}
where $\gamma \colon (0,1) \to \R_\ge$ is given by $\gamma(u):= \partial_{-} g(x)|_{x = 1 - u}$, $\partial_{-}$ denotes the derivative from the left, the constant $\Lambda:= \left( \lambda + \mu\,c[X^\theta,X^\phi]\right)\mathds{1}_{d_p[X^\theta,X^\phi]>\ep}$, and $f_\theta(\cdot)$ is the density of $X^\theta$.
\end{proposition}

The gradient formula \eqref{eqn:L-gradient-inner} requires estimating the function $\nabla_\theta F_\theta(x)$. For this purpose, suppose we are given a mini-batch of data $\{(x_\phi^{(1)},x_\theta^{(1)}),\dots,(x_\phi^{(N)},x_\theta^{(N)})\}$ of $(X^\phi,X^\theta)$, which, e.g., may be the result of an accumulation of multiple sources of randomness (such as in dynamic trading). We then make a kernel density estimator (KDE) $\hat{F}_\theta$ of $F_\theta$ given by
\begin{equation}
    \hat{F}_\theta(x) = \tfrac{1}{N}\sum_{i=1}^N \Phi\big(x- x^{(i)}_\theta\big),
\end{equation}
where $\Phi(\cdot)$ denotes the distribution function for an appropriate (zero-centred and standardised) kernel (e.g., Gaussian). Gaussian kernels are generally quite flexible, but other choices  (such as Epanechnikov, Triweight, and Triangular) are possible, for discussions on kernel and bandwidth selection more generally see, e.g., \cite{gramacki2018nonparametric}. The kernel choice made little difference in our experiments.
Therefore, 
\begin{equation}
    \nabla_\theta \hat{F}_\theta(x) = -\tfrac{1}{N}\sum_{i=1}^N \Phi^\prime\big(x-x^{(i)}_\theta\big) \,\nabla_\theta x^{(i)}_\theta,
\end{equation}
where $\Phi^\prime(\cdot)$ is the  kernel's corresponding density. As the samples $x_\theta^{(i)}$ are viewed as outputs of an ANN, the gradients $\nabla_\theta x_\theta^{(i)}$ may be efficiently obtained using standard back-propagation techniques.

Inserting the KDE into \eqref{eqn:L-gradient-inner}, we may estimate the gradient by
{
\begin{multline}
    \nabla_\theta L[\theta,\phi] \approx  - \frac{1}{N}
    \sum_{i=1}^N\left[ \Bigg( U'(x_\theta^{(i)})\gamma\left(\hat{F}_\theta(x_\theta^{(i)})\right)
    \right.
    \\
    \left.
    - p\,\Lambda\,
     |x_\theta^{(i)} - x_{\phi,c}^{(i)}|^{p-1}
     \sgn(x_\theta^{(i)} - x_{\phi,c}^{(i)})\Bigg)
     \tfrac{
     \sum_{j=1}^N \Phi^\prime(x_\theta^{(i)}-x_\theta^{(j)} ) \,\nabla_\theta x_\theta^{(j)}
     }
     {
     \sum_{k=1}^N \Phi^\prime(x_\theta^{(i)}-x_\theta^{(k)} )
     }
     \right]\,,
     \label{eqn:inner-gradient-sim}
\end{multline}
}
where $x_{\phi,c}^{(1)}, \ldots x_{\phi,c}^{(N)}$ are the reordered realisations of $X^\phi$, such that they are comonotonic with $X^\theta$. The Wasserstein distance between $X^\theta$ and $X^\phi$ may be approximated using the same mini-batch as $\left(\frac1N \sum_{i = 1}^N |x_\theta^{(i)} - x_{\phi, c}^{(i)}|^p\right)^\frac1p$, see e.g., \cite{ambrosio2003lecture}[Chapter 1.].

\subsection{The Outer Problem}

Similar to the inner problem, optimisation for the outer problem is carried out using the augmented Lagrangian, this time taking gradients wrt $\phi$. To calculate the derivatives, we must specify how $X^\theta$ is generated. Specifically, we assume $X^\theta=H_\theta(X^\phi,Y)$ where $Y$ is another (multi-dimensional) source of randomness. 

\begin{proposition}[Outer  Gradient Formula.]
\label{prop:outer-gradient-formula}
Let $X_c^\phi$ denote the version of $X^\phi$ which makes $(X^\theta,X_c^\phi)$ comonotonic -- i.e., reorder the realisations of $X_c^\phi$ according to the rank of $X^\theta$ -- then the gradient becomes
\begin{multline}
    \nabla_\phi L[\theta,\phi] =\E\left[ U^\prime(X^\theta)\,\gamma(F_\theta(X^\theta)) 
    \frac{\nabla_\phi F_\theta(x)|_{x=X^\theta}}{f_\theta(X^\theta)}
    \right.
    \\
    \left.
    -p\,\Lambda\,
     |X^\theta - X_c^\phi|^{p-1} \sgn(X^\theta - X_c^\phi)
      \left(\frac{\nabla_\phi F_\theta(x)|_{x=X^\theta}}{f_\theta(X^\theta)}+ \frac{\nabla_\phi G_\phi(x)|_{x=X^\phi}}{g_\phi(X^\phi)}\right)
     \right],
     \label{eqn:L-gradient-outer}
\end{multline}
where the constant $\Lambda:= (\lambda 
    + \mu\,c[X^\theta,X^\phi]^p)\mathds{1}_{d_p[X^\theta,X^\phi]>\ep}$ and $g_\phi(\cdot)$ is the density of $X^\phi$.
\end{proposition}
As in the previous section,  given a mini-batch $\{(x_\phi^{(1)},x_\theta^{(1)},y^{(1)}),\dots,(x_\phi^{(N)},x_\theta^{(N)},y^{(N)})\}$ of \linebreak $(X^\phi,X^\theta,Y)$, we may estimate the gradient by
\begin{multline}
    \nabla_\phi L[\theta,\phi] \approx  -\frac{1}{N}    \sum_{i=1}^N\left[   
     U^\prime(x_\theta^{(i)}) \gamma\left(\hat{F}_\theta(x_\theta^{(i)})\right)
\tfrac{
     \sum_{j=1}^N \Phi^\prime(x_\theta^{(i)}-x_\theta^{(j)} ) \,\nabla_\phi x_\theta^{(j)}
     }
     {
     \sum_{k=1}^N \Phi^\prime(x_\theta^{(i)}-x_\theta^{(k)} )
     }     
     \right.
     \\
     \left.
     -p\,\Lambda
|x_\theta^{(i)} - x_{\phi,c}^{(i)}|^{p-1}\sgn(x_\theta^{(i)} - x_{\phi,c}^{(i)})\,
     \left(\tfrac{
     \sum_{j=1}^N \Phi^\prime(x_\theta^{(i)}-x_\theta^{(j)} ) \,\nabla_\phi x_\theta^{(j)}
     }
     {
     \sum_{k=1}^N \Phi^\prime(x_\theta^{(i)}-x_\theta^{(k)} )
     } + 
     \tfrac{
     \sum_{j=1}^N \Phi^\prime(x_{\phi,c}^{(i)}-x_{\phi,c}^{(j)} ) \,\nabla_\phi x_{\phi,c}^{(j)}
     }
     {
     \sum_{k=1}^N \Phi^\prime(x_{\phi,c}^{(i)}-x_{\phi,c}^{(k)} )
     }\right)
     \right]\,,
     \label{eqn:outer-gradient-sim}
\end{multline}
where we use for simplicity the same kernel for $F_\theta$ and $G_\phi$. The gradients $\nabla_\phi x_{\phi,c}^{(j)}$ and  $\nabla_\phi x_\theta^{(j)}$ may be computed using the relationship $x_\theta^{(j)}=H_\theta(x^{(j)}_\phi,y^{(j)})$ and back-propagation.
Algorithm \ref{alg:generic} provides an overview of the optimisation methodology.

\subsection{Randomised Policies}\label{sub-sec-random-action}

To explore the state-space better, randomised (also known as probabilistic) policies  are often used to achieve the so-called exploration/exploitation trade off \cite{wang2020reinforcement,firoozi2020exploratory,guo2020entropy}. There are also contexts, such as robot control, where one can only control the policy distribution. As such, we briefly discuss how the results in Propositions \ref{prop:inner-gradient-formula} and \ref{prop:outer-gradient-formula} may be applied in the randomised  policy case. For example, it is often the case
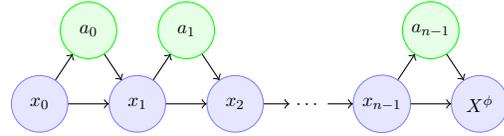
\begin{wrapfigure}{r}{0.5\textwidth}
\centering
\begin{tikzpicture}[scale=0.65,every node/.style={transform shape}]
\node[obs, minimum size=3em] (x0) at (-2,0) {$x_{0}$};
\node[action, minimum size=3em] (a0) at (-1,1.5) {$a_0$};

\node[obs, minimum size=3em] (x1) at (0,0) {$x_1$};
\node[action, minimum size=3em] (a1) at (1,1.5) {$a_1$};

\node[obs, minimum size=3em] (x2) at (2,0) {$x_2$};

\node (x3) at (3.5,0) {$\dots$};
\draw [->] (x2) to (x3);

\node [obs, minimum size=3em] (x4) at (5,0) {$x_{n-1}$};
\node [action, minimum size=3em] (a4) at (6,1.5) {$a_{n-1}$};
\draw [->] (x3) to (x4);

\node [obs, minimum size=3em] (x5) at (7,0) {$X^\phi$};

\draw [->] (x0) to (a0);
\draw [->] (a0) to (x1);
\draw [->] (x0) to (x1);

\draw [->] (x1) to (a1);
\draw [->] (a1) to (x2);
\draw [->] (x1) to (x2);

\draw [->] (x4) to (a4);
\draw [->] (a4) to (x5);
\draw [->] (x4) to (x5);
\end{tikzpicture}
\caption{Graphical model representation of randomised policies. \label{fig:random-policies}}
\end{wrapfigure}
that the terminal rv $X^\phi$ (of the outer problem) stems from a sequence of actions $a_{0:n-1}$  that are conditionally generated from the previous system states $x_{0:n}$, where $x_n = X^\theta$, as in the graphical model in Figure \ref{fig:random-policies}. Hence, the probability density function (pdf) over the sequence of state/action pairs admits the decomposition
\begin{equation}
    g_\phi(x_{0:n},a_{0:n-1}) = h(x_0)\prod_{t=0}^{n-1} \pi_\phi(a_t|x_t)\, h(x_{t+1}|x_t,a_t),
\end{equation}
where $h(x_{t+1}|x_t,a_t)$ specifies the conditional one-step transition densities, $h(x_0)$ is the prior on $x_0$, and $\pi_\phi(a_t|x_t)$ the pdf of actions conditioned on states. To compute the gradient $\nabla_\phi G_\phi(x)$ we use the above decomposition and note that
\begin{equation}
    G_{\phi}(x)
    = \int_{\R}\dots\int_{\R}\int_{-\infty}^x h(x_0)\prod_{t=0}^{n-1} \pi_\phi(a_t|x_t)\, h(x_{t+1}|x_t,a_t)\;dx_0\dots dx_{n-1}\, dx_n,
\end{equation}
Therefore, its gradient becomes
{\footnotesize
\begin{align*}
    \nabla_\phi G_\phi(x)  
    &= 
    \int_{\R}\dots\int_{\R}\int_{-\infty}^x\sum_{t'=0}^{n-1} \nabla_\phi \pi_\phi(a_{t'}|x_{t'})\;
    h(x_0)\prod_{\substack{t=0\\t\ne t'}}^{n-1} \pi_\phi(a_t|x_t)\, h(x_{t+1}|x_t,a_t)\;dx_0\dots dx_{n-1}\, dx_n
    \\
    \;&= 
    \int_{\R}\dots\int_{\R}\int_{-\infty}^x
    \left(\sum_{t'=0}^{n-1} \nabla_\phi \log \pi_\phi(a_{t'}|x_{t'})\right)h(x_0)\prod_{t=0}^{n-1} \pi_\phi(a_t|x_t)\, h(x_{t+1}|x_t,a_t)\;dx_0\dots dx_{n-1}\, dx_n
    \\
    \;&=
    \E\left[\sum_{t=0}^{n-1} \nabla_\phi \log \pi_\phi(a_{t}|x_{t})\;\Id_{\{X^\phi\le x\}}\right].
\end{align*}
}%
In the last line, $(a_t,x_t)_{t=0,\dots,n}$ should be understood as rvs corresponding to the outputs of all nodes in the graphical model in Figure \ref{fig:random-policies}.
Thus, using a KDE approximation from samples of state-action sequences $\{(x^{(m)}_0,a_0^{(m)},\dots, x^{(m)}_{n-1},
a_{n-1}^{(m)},x^{(m)}_n)_{m=1,\dots N}\}$, we may estimate the gradient by
\begin{equation}
    \nabla_\phi \hat{G}_\phi(x)
    \approx
    \tfrac{1}{N}\sum_{m=1}^N \sum_{t=0}^{n-1} \nabla_\phi \log \pi_\phi(a_{t}^{(m)}|x_{t}^{(m)})\, \Phi(x^{(m)}_n-x).
\end{equation}
To obtain a more explicit form, one must specify how actions are drawn using a given policy, e.g., they may be normally distributed with mean and standard deviation parameterised by an ANN. The remaining gradient $\nabla_\phi \log \pi_\phi(a|x)$ may then be computed using back-propagation along the sampled mini-batch of paths. Similar calculations can be performed to derive formulae for $\nabla_\phi F_\theta$, note that $\nabla_\theta F_\theta$ does not have a gradient wrt actions.

\section{Examples}\label{sec:example}
Here, we illustrate the three prototypical examples described earlier. For this, the investor's RDEU is a combination of a linear utility and an $\alpha$-$\beta$ distortion given by:
\begin{equation}\label{eqn:gamma}
    \gamma(u) = \tfrac{1}{\eta}\left( \mfp\,\Id_{\{u\le \alpha\}} + (1-\mfp)\,\Id_{\{u>\beta\}}\right),
\end{equation}
with normalising constant $\eta=\mfp\,\alpha+(1-\mfp)\,(1-\beta)$, $0<\alpha\le\beta<1$, and $\mfp\in[0,1]$. This 
\begin{wrapfigure}{r}{0.52\textwidth}
\begin{minipage}{0.52\textwidth}
\begin{algorithm}[H]
\tiny
\SetAlgoLined
initialise networks $\theta,\phi$;

initialise Lagrangian multipliers $\lambda = 1$, $\mu = 10$, $\alpha = 1.5$;

\For{$i\leftarrow 1$ \KwTo $M_{outer}$}
{
Simulate mini-batch of $X^\phi$;

\For{$j\leftarrow 1$ \KwTo $M_{inner}$}
{

Simulate mini-batch of $X^\theta$ using fixed $X^\phi$ in outer loop;

Estimate inner gradient $\nabla_\theta L[\theta,\phi]$ using \eqref{eqn:inner-gradient-sim};

Update network $\theta$ using a ADAM step;

\If{$ (j + 1) \% N_{Lagrange} = 0 $}{
    Update multipliers: $\lambda \leftarrow\lambda + \mu\,c(\theta^*)$ and $\mu\leftarrow \alpha\,\mu $;
}

Repeat until $d_p[X^\theta,X^\phi]\le \ep$, and $\RM^U_{\gamma}[X^\theta]$ has not increased for the past 100 iterations;

}

Simulate mini-batch of $X^\theta$ from $X^\phi$ and trained $\theta$ network;

Estimate outer gradient $\nabla_\phi L[\theta,\phi]$ using \eqref{eqn:outer-gradient-sim};

Update network $\phi$ using a ADAM step;

Repeat until $\RM^U_{\gamma}[X^\theta]$ has not decreased for the past 100 iterations;
}
\caption{Schematic of optimisation algorithm.\label{alg:generic}}
\end{algorithm}
\end{minipage}
\end{wrapfigure}
parametric family is $U$-shaped (i.e., $S$-shaped RDEU), which is well-known to account for the investor's loss avoiding while simultaneously risk-seeking behaviour, and contains several notable risk measures as special cases. For $\mfp = 1$, it reduces to the CVaR at level $\alpha$, for $\mfp= 0$, to the upper tail expectation (UTE) at level $\beta$, and for $\mfp>\frac{1}{2}$ ($\mfp<\frac{1}{2}$), it emphasises losses (gains) relative to gains (losses). For all experiments, unless otherwise stated, we use $\alpha=0.1$, $\beta=0.9$, and $\mfp=0.75$ to showcase how investors protect themselves from downside risk while still seeking gains.

In the examples below, before computing the outer gradient, we ensure that constraints of the inner problem are binding, so that $\Lambda = 0$ in \eqref{eqn:L-gradient-outer}. Furthermore, while it is easy to incorporate transaction costs in all of these examples, we opted to exclude them for simplicity of the settings. Convergence for the inner loop is said to be achieved when $\RM[X^\theta]$  changes by less than $1\%$ (or 5000 iterations reached), and the outer loop  when $\RM[X^\phi]$ changes by less than $1\%$ (or 500 iterations reached). Each iteration of the outer loop (which includes convergence of the inner loop) of Examples 1 and 3 executed in  $79.8\pm0.2s$ and $62.7\pm0.3s$, respectively, and convergence was achieved in about 66min and 52min, respectively. Each inner loop for Example 2 executed in $8.1\pm0.1s$ and convergence was achieved in about $66.6$min. All using a Tesla P100 GPU.

\subsection{Robust Portfolio Allocation} In this subsection, we illustrate the results on a problem introduced in Example \ref{ex:Robust-Portfolio-Allocation}. We take the setup from \cite{esfahani2018data} where the market consists of $d$-assets whose returns are driven by a systematic factor $\zeta\sim\mathcal{N}(0, 0.02^2)$ and idiosyncratic factors 
$Z_i\sim\mathcal{N}(0.03\,i,0.025^2\,i^2)$, $i \in \D:=\{1,\dots,d\}$, where the factors $\zeta,Z_1,\dots,Z_d$ are mutually independent.
The individual returns are $R_i=\zeta+Z_i$ and the total return is $X^\phi=\phi^\intercal R$.
Such a model can easily be generalised to include several systemic factors. We model the outer strategy $\pi^\phi$ as an ANN that maps a zero tensor directly to the asset weights $\phi$ through a softmax activation function (to avoid short-selling: $\phi_i\ge0$, $i\in\D$, $\sum_{i\in\D}\phi_i=1$) of the learned bias. We use the Wasserstein distance of order $p=1$.

\begin{figure}[t]
    \centering
    \includegraphics[align=c,height=0.25\textheight]{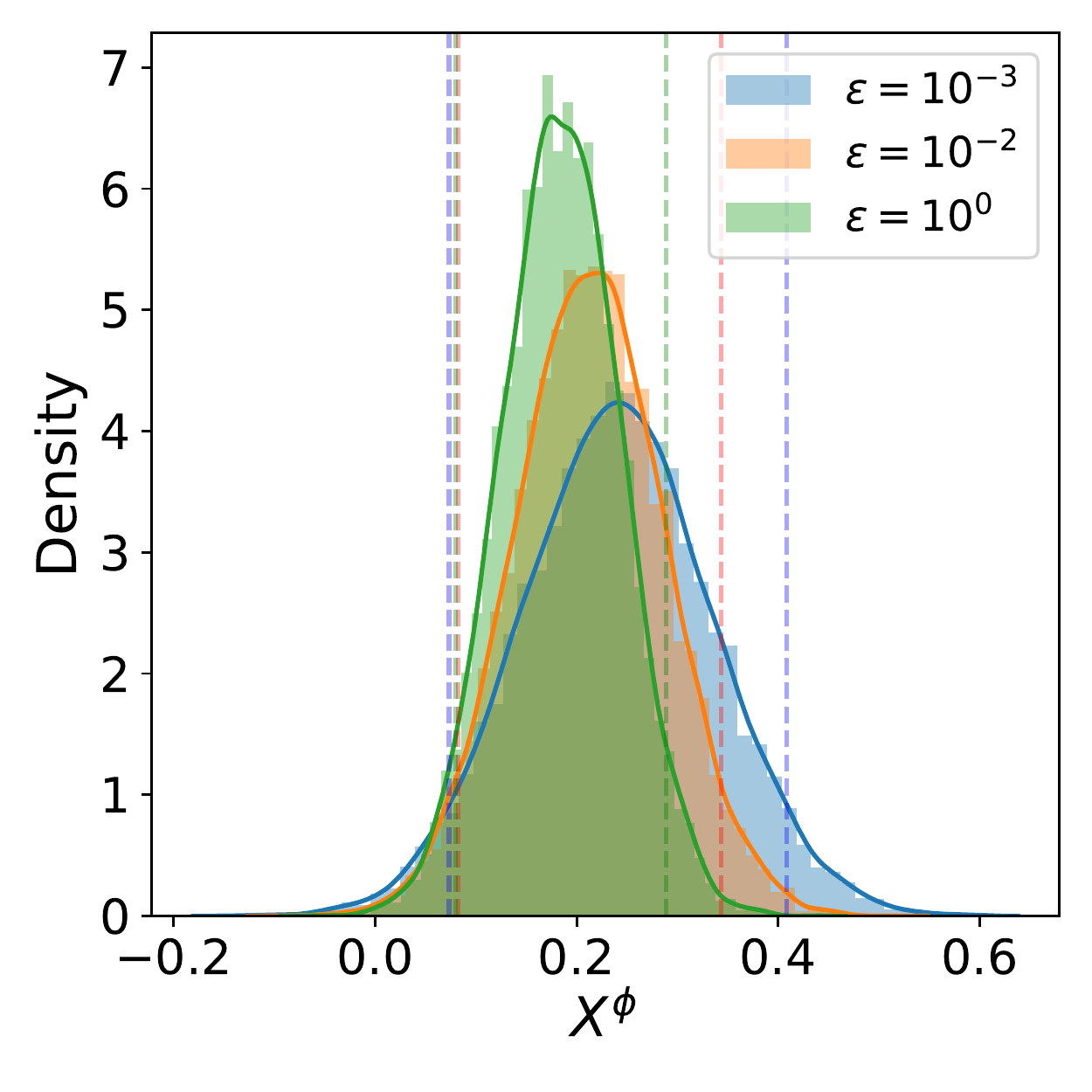}
    \quad
    \includegraphics[align=c,height=0.25\textheight]{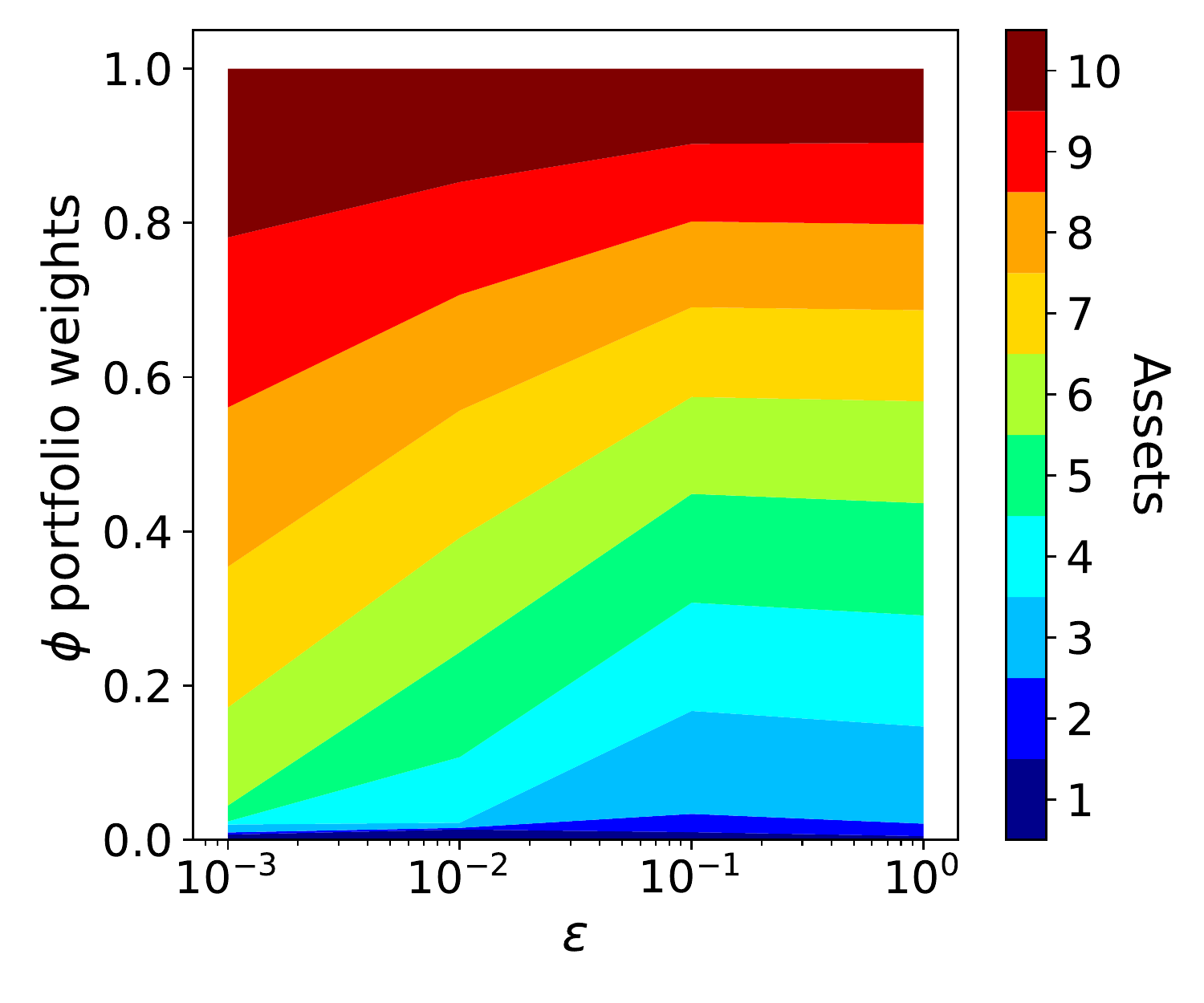}
    \caption{Robust portfolio allocations as $\ep$ varies. (Left) densities of terminal wealth, and (right) percentage of wealth held in each asset. Dashed vertical lines: $CVaR_\alpha$ and $UTE_\beta$.}
    \label{fig:RobustPortfolioAllocation}
\end{figure}
Figure \ref{fig:RobustPortfolioAllocation} illustrates the optimal terminal wealth (left panel) and percentage of investment as a function of the size of the Wasserstein ball $\ep$ (right panel), for $d=10$ assets. For larger $\ep$, the investor seeks more robustness, which is illustrated in the left panel that shows that all $CVaR_\alpha$ ($UTE_\beta$) move to the left as $\ep$ increases. Specifically, for $\ep$ equal to $10^{-3}$,$10^{-2}$, and $10^0$, we have $CVaR_\alpha=$ $0.073$, $0.082$, and $0.08$ and $UTE_\beta=$ ($0.41$, $0.34$, $0.29$). These statistics indicate that the investor becomes more and more conservative with increasing Wasserstein distance. This is further emphasised in the right panel, where, for small $\ep$, the investor puts most of their wealth in the riskier assets, and as $\ep$ increases, moves closer to an equally weighted portfolio. 
For completeness, the $CVaR_\alpha$ ($UTE_\beta$) of the worst-case distribution around the optimal $X^\phi$, as $\ep$ varies from $10^{-3}$,$10^{-2}$, $10^0$ are $0.071$, $-0.012$, and $-9.942$ ($0.401$, $0.341$, $0.285$), respectively.

\subsection{Optimising Risk-Measures with a Benchmark}

Next, we illustrate our RL approach on a portfolio allocation problem where an agent aims to improve upon a benchmark strategy, as described in Example \ref{ex:benchmark}. The outer problem has a singleton corresponding to a benchmark strategy $\phi$ -- which we take to be a constant proportion of wealth strategy (any other benchmark strategy would do) -- and the agent aims to seek over alternate strategies that minimise RDEU, i.e. replacing sup with inf in the inner problem.

We optimise in discrete time $\mathcal{T}:=\{0,1,\dots,T=5\times252\}$ and consider an investor who chooses from the set of admissible strategies $\A$ consisting of $(\F_t)_{t\in\mathcal{T}}$-adapted Markov processes that are self-financing and in $\Lp(\Omega,[0,T])$. For an arbitrary $\pi_\theta\in\A$, where $\pi_\theta:=((\pi_{\theta,t}^{i})_{i\in\D})_{t\in\mathcal{T}}$, $\D:=\{1,\dots,d=3\}$, represents the percentage of wealth invested in each asset, the investor's wealth process  $X^{\pi_\theta}:=(X^{\pi_\theta}_t)_{t\in\mathcal{T}}$ satisfies the usual self-financing equation. To illustrate the flexibility of our formulation, we use a stochastic interest rate model combined with a constant elasticity of variance (SIR-CEV) market model. The details of the market model dynamics and parameters may be found in \cite{pesenti2020portfolio}. Here, we use the Wasserstein distance of order $p=2$.

\begin{figure}[t]
\centering
    \includegraphics[height=0.25\textheight]{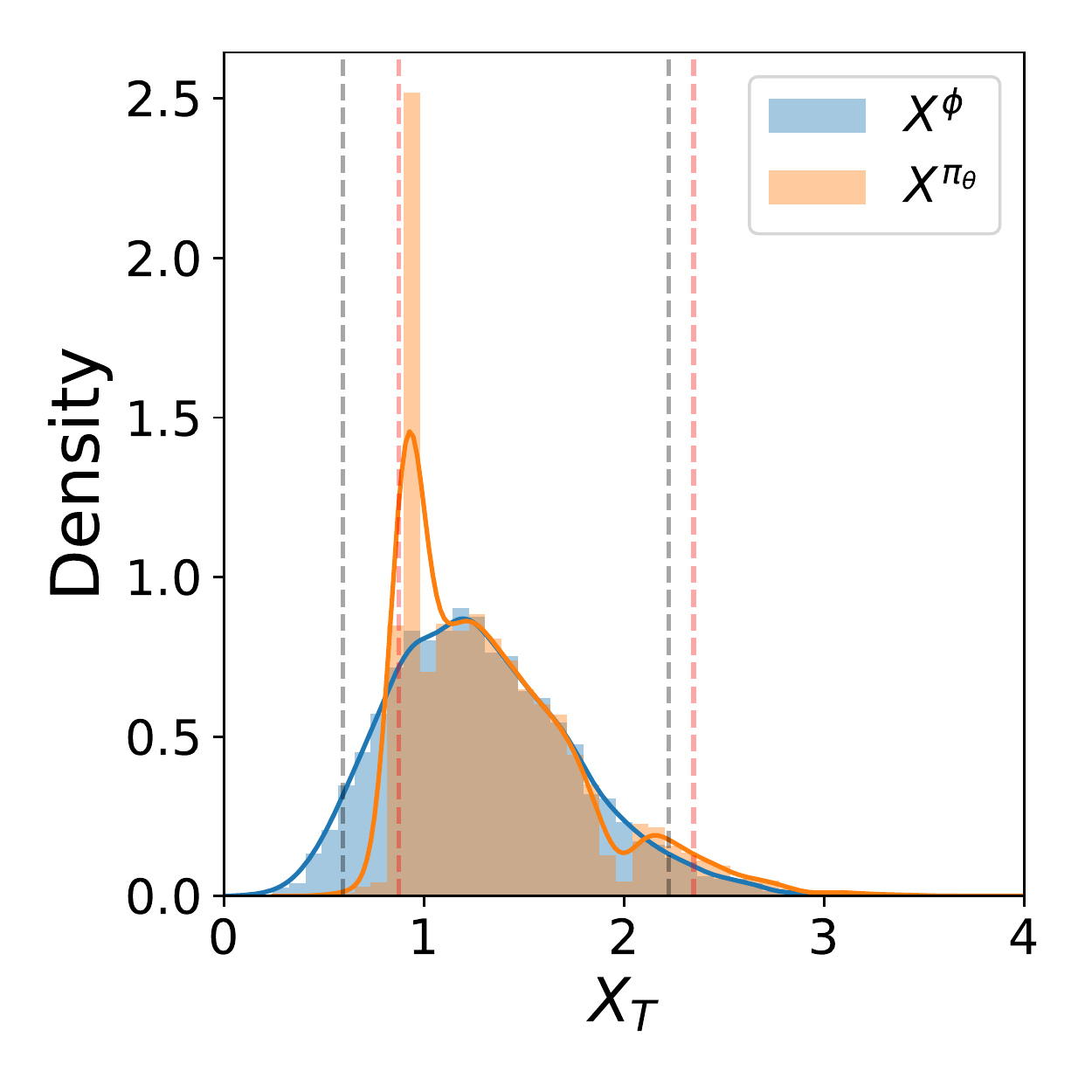}
    \quad
    \includegraphics[height=0.25\textheight]{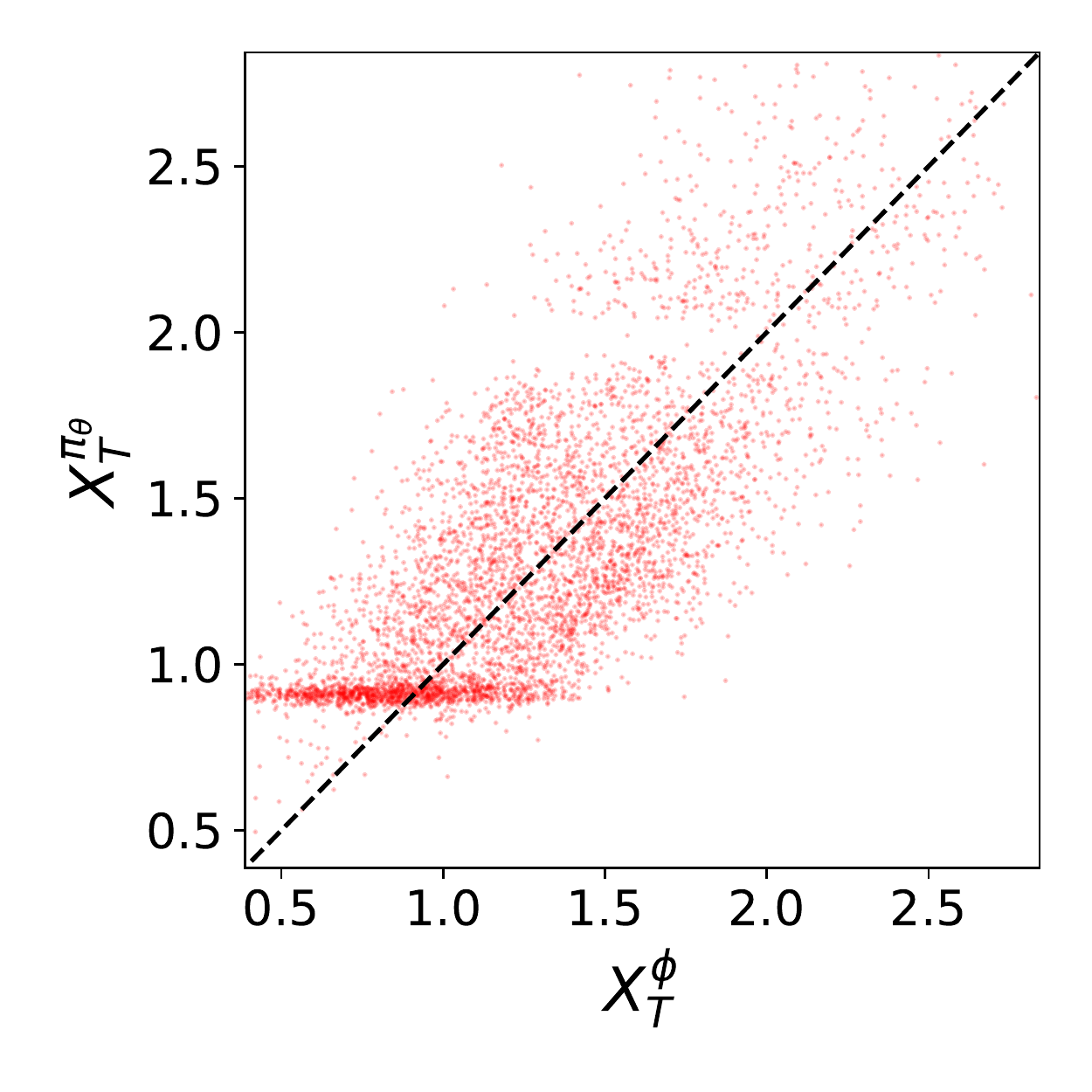}
    \caption{Illustrations of the terminal wealth rv of the benchmark's $X_T^\phi$ and of the optimal's $X_T^{\pi_\theta}$. Vertical lines indicate the $CVaR_{\alpha}$ and $UTE_\beta$.}
    \label{fig:benchmark}
\end{figure}
We use a fully connected feed-forward neural network with 3 hidden layers of 50 neurons each with ReLU activation functions to map the state consisting of the features $\{t, S_t, X_t^{\phi}\}$ to 
policy
$\pi_{\theta,t}$.
The final output layer may be chosen to reflect constraints on the portfolio weights -- e.g., long only weights with no leveraging, in which case one would use a sigmoid activation
function. 
For the results shown here we  have no activation for the final layer and thus allow the agent to take long, short, and leveraged positions.

Figure \ref{fig:benchmark} illustrates the resulting optimal portfolio $\pi_\theta$ for a constant proportion benchmark strategy $\phi$. The left panel shows the density of the benchmark $X_T^\phi$ and optimal $X_T^{\pi_\theta}$ terminal wealth. The optimal pulls mass from the left tails into a spike near $VaR_\alpha$ and pushes mass into the right tail. This reflects the investor's risk preferences. The right scatter plot shows the state-by-state dependence between the optimal and the benchmark. The results are qualitatively similar to those derived in \cite{pesenti2020portfolio}, even though here the problem is posed in discrete time. While the analytical approach in \cite{pesenti2020portfolio} applies only for complete market models, the proposed RL approach is also applicable to incomplete markets.

\subsection{Robust Statistical Arbitrage}
\label{sec:RobustStatArb_example}
In this subsection, we explore Example \ref{ex:dynamic}. In this case, 
the outer strategy $\pi^\phi=(\pi^\phi_t)_{t\in\mathcal{T}}$ denotes the position the trader holds at discrete times $\mathcal{T}=\{0,\Delta T,\dots,N\Delta T\}$. 
We assume the asset price process $S=(S_t)$ is an Ornstein-Uhlenbeck process satisfying $dS_t=(\kappa(b-S_t)+c\sgn{\phi_{j_t}}\sqrt{|\phi_{j_t}|})\,dt+\sigma\,dW_t$ with $\kappa=5$, $b=1$, $\sigma=0.8$, $N=252$, $c=0.1$, $\Delta T=\frac{1}{252}$, and $j_t=	\lfloor\frac{t}{\Delta t}\rfloor$. 
The outer strategy $\pi^\phi$ is determined by a fully connected feed-forward neural network with 3 hidden layers of 50 neurons each with ReLU activation functions, and takes the state consisting of the features $\{t, S_t, q_{t-1}^\phi\}$ as input (recall $q_t^\phi=\sum_{j=0}^{t-1}\phi_j$ is the controlled inventory). The final output layer is chosen to reflect the constraints on the inventory. Here, we use a $5\tanh$ activation function  to constrain the strategy such that inventory remains in the interval $[-5,5]$.  We use the Wasserstein distance of order $p=2$.

\begin{figure}[t]
\centering
\includegraphics[height=0.25\textheight]{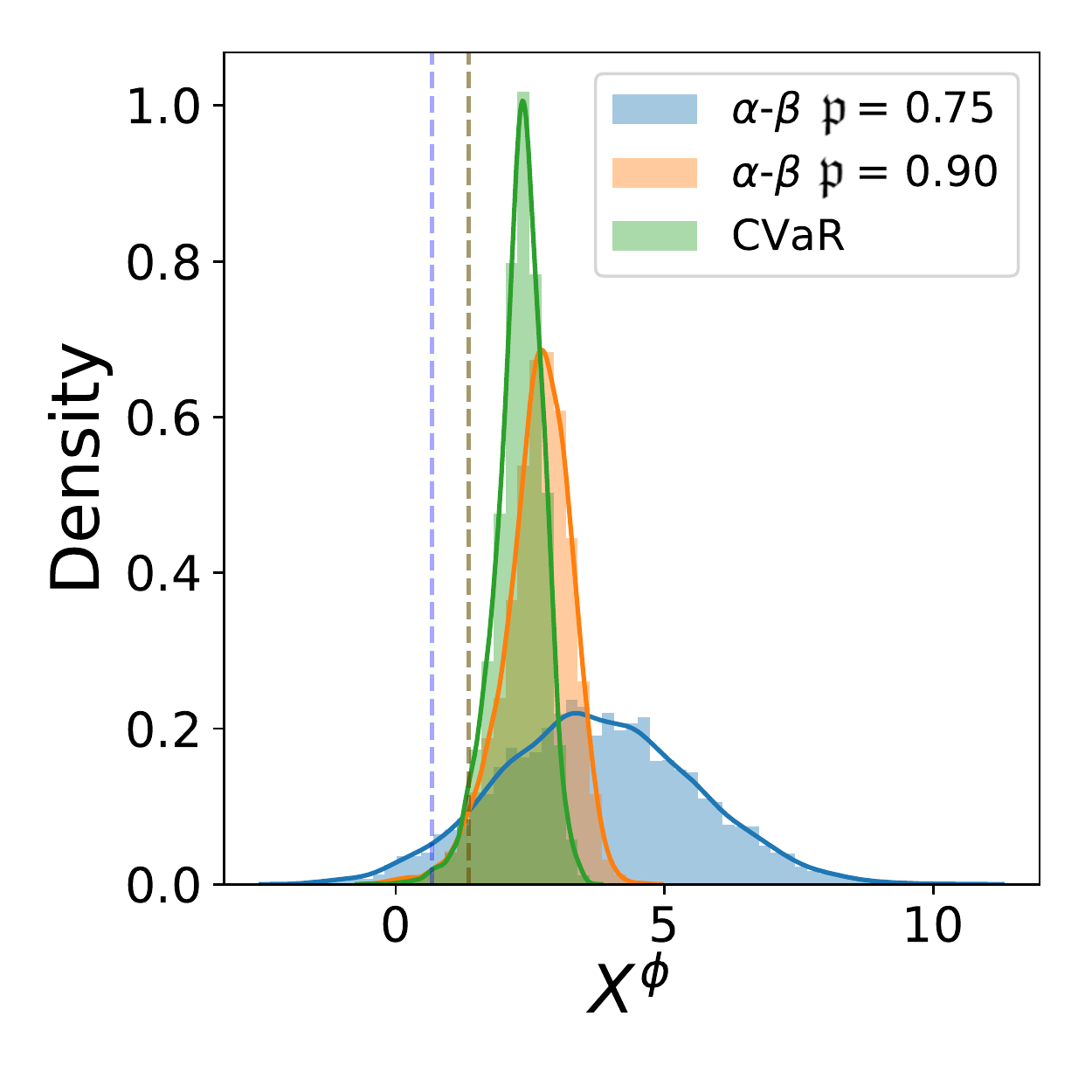}
\quad
\includegraphics[height=0.25\textheight]{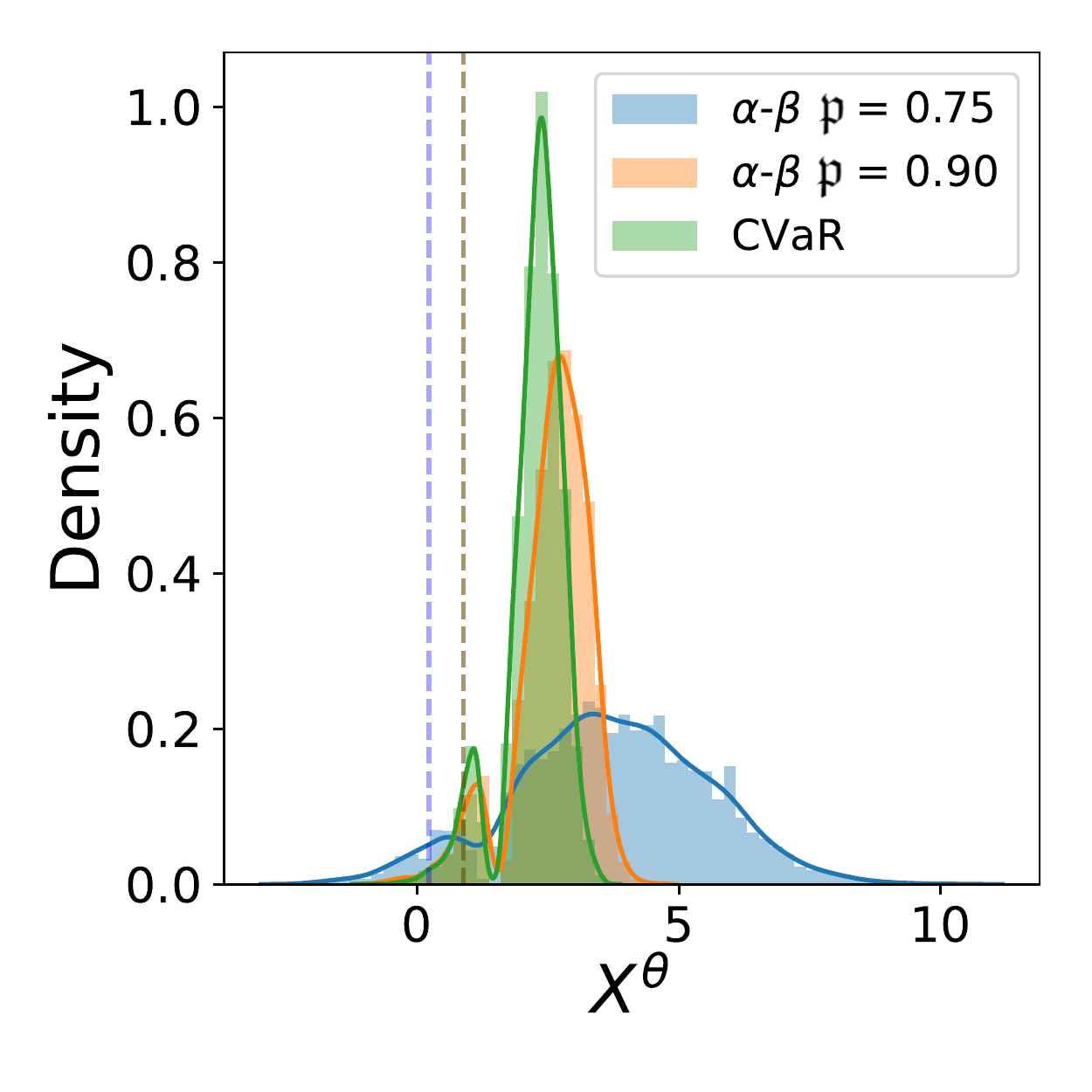}
\caption{
Optimal statistical arbitrage strategy's terminal densities (left) and corresponding worst-case densities (right). Vertical dashed lines show the corresponding $CVaR_{0.1}$.\label{fig:stat-arb-}
}
\end{figure}
\begin{figure}[t]
    \centering
    \begin{minipage}[t]{0.18\textheight}
    \centering
    \includegraphics[width=\textwidth]{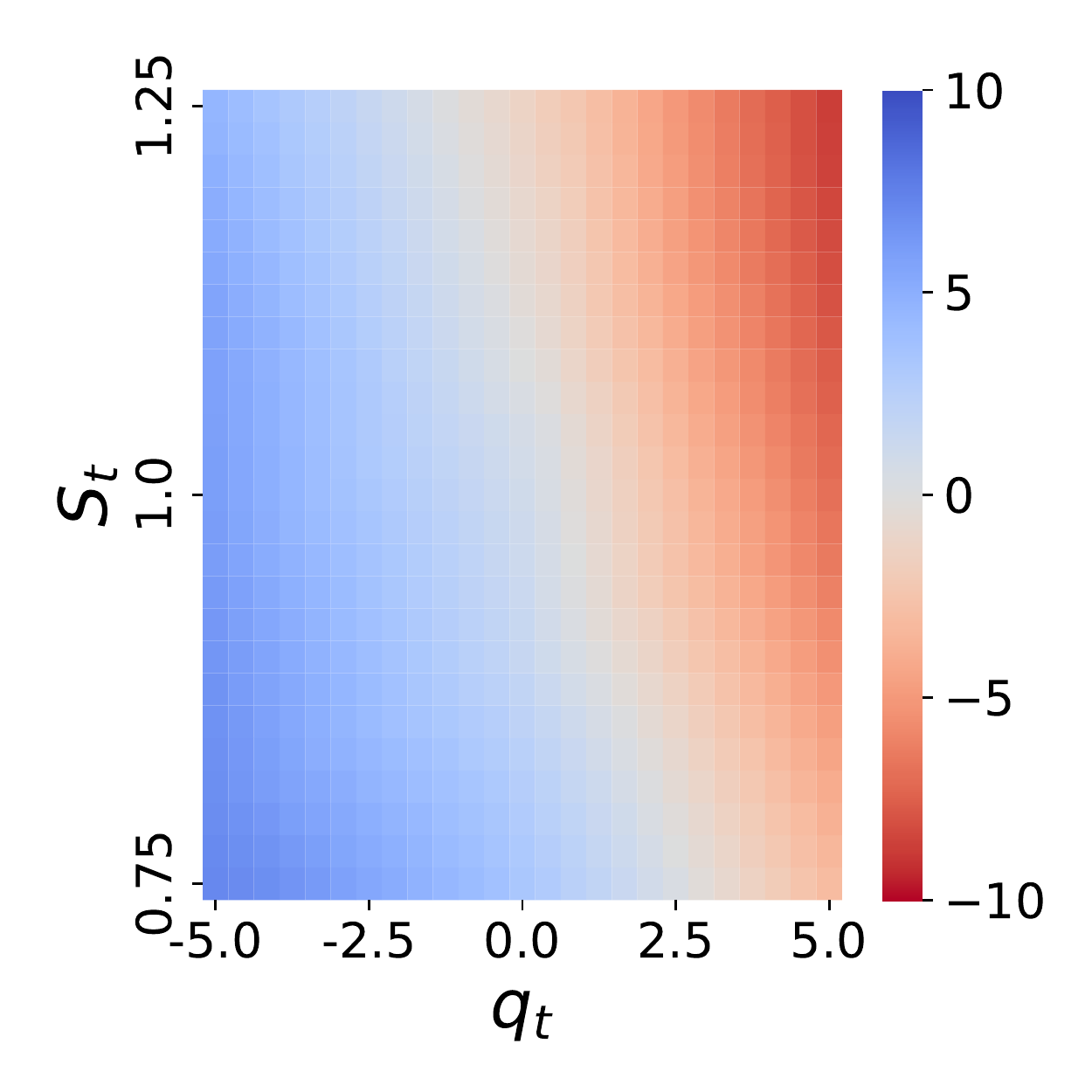}
    \\[-0.5em]
    $\mfp=1$
    \end{minipage}
    \begin{minipage}[t]{0.18\textheight}
    \centering
    \includegraphics[width=\textwidth]{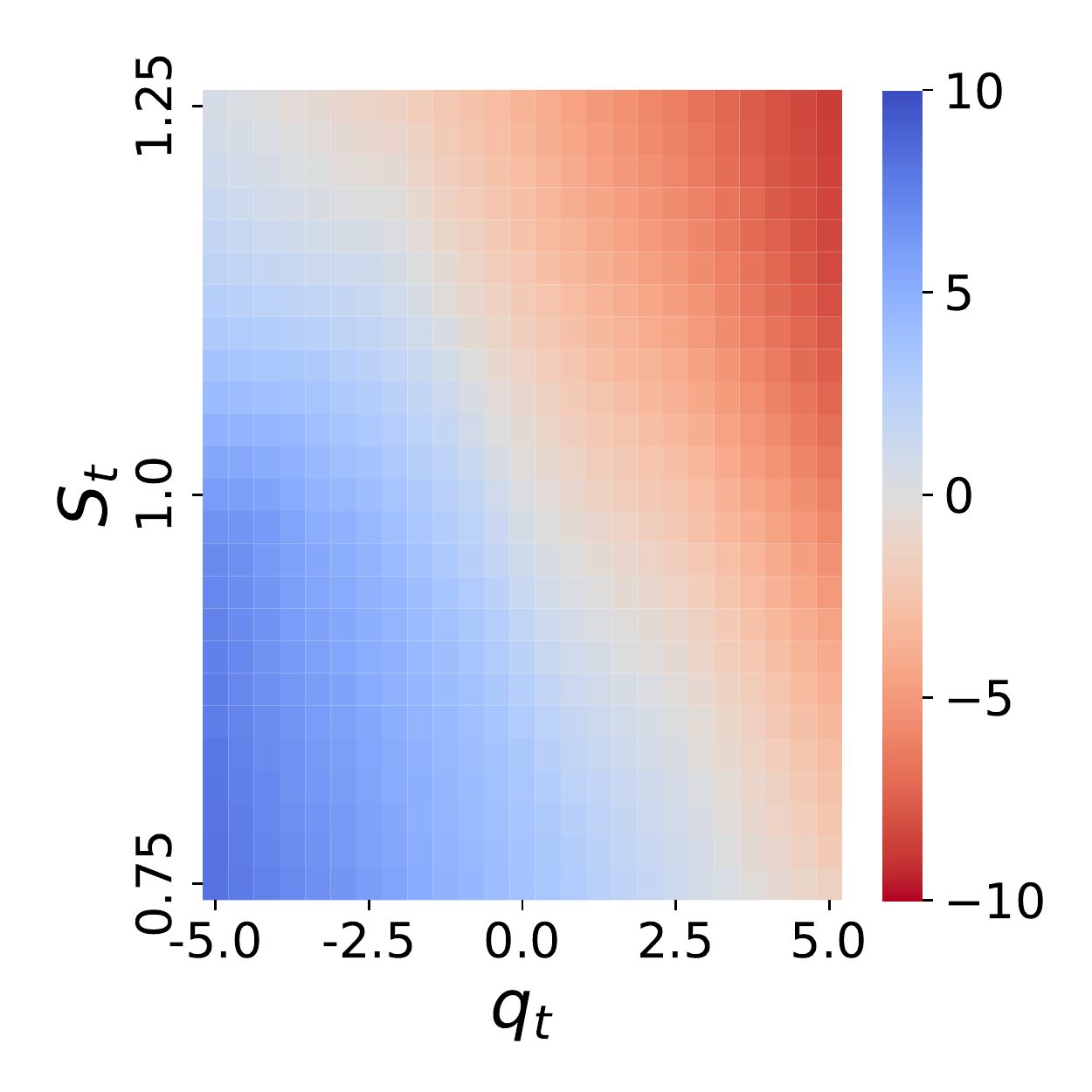}
    \\[-0.5em]
    $\mfp=0.9$
    \end{minipage}
    \begin{minipage}[t]{0.18\textheight}
    \centering
    \includegraphics[width=\textwidth]{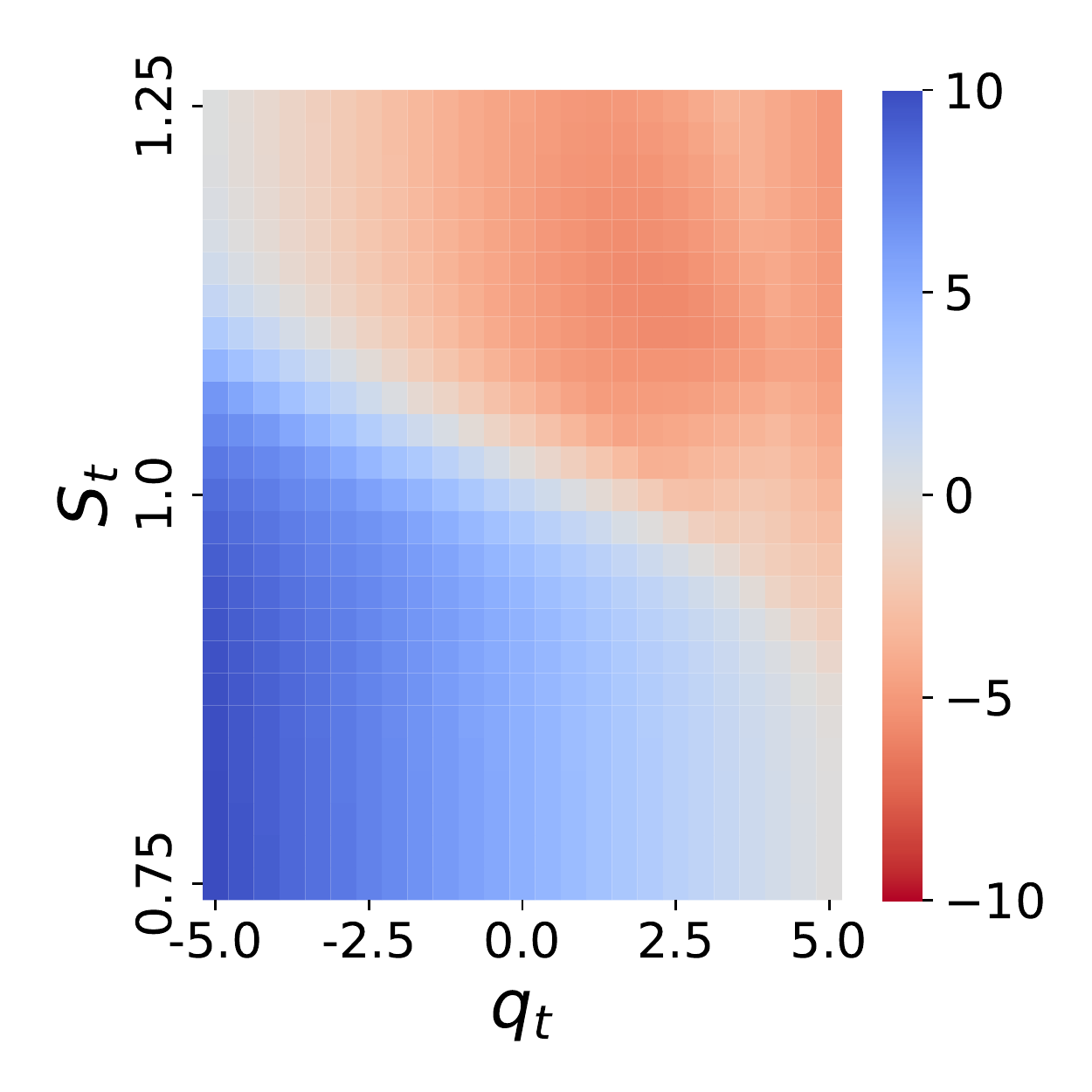}
    \\[-0.5em]
    $\mfp=0.75$
    \end{minipage}
    \begin{minipage}[t]{0.18\textheight}
    \centering
    \includegraphics[width=\textwidth]{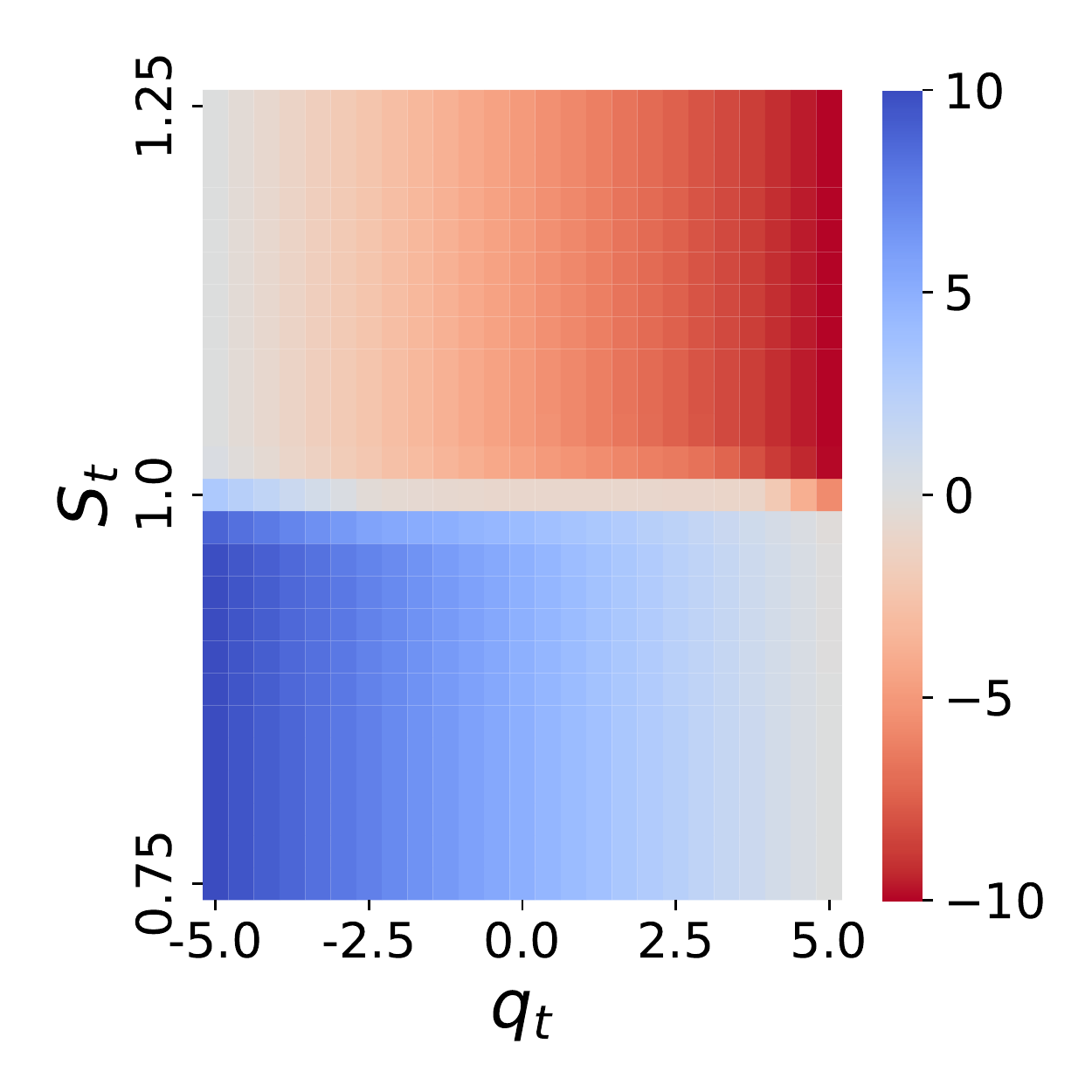}
    \\[-0.5em]
    risk-neutral
    \end{minipage}    
    \caption{Optimal statistical arbitrage strategies ($t=0.75\,T$) for risk-neutral and $\alpha$-$\beta$ risk measures.}
    \label{fig:stat-arb-optimal-strategy}
\end{figure}

Figure \ref{fig:stat-arb-} show results for the $\alpha$-$\beta$ risk measure for $\mfp=1$ ($CVaR_\alpha$), $\mfp=0.9$, and $\mfp=0.75$. The left panel shows the optimal densities as $\mfp$ varies and the right panel the corresponding worst-case densities that result from solving the inner problem with the optimal as the reference distribution. For increasing $\mfp$, the agent puts more and more weight on the upper tail and the optimal distribution becomes more profitable, but also more risky (the $CVaR_{0.1}$ moves to the left). In Figure \ref{fig:stat-arb-optimal-strategy}, we illustrate the optimal execution strategies at time $t = 0.75\,T$ through a heat map and as a function of current inventory and asset price. The colours indicate the optimal trade -- e.g., a location in  deep red indicates short selling of $10$ units of the asset. As $\mfp$ decreases the agent becomes more gain-seeking and starts taking more aggressive actions to take advantage of the mean-reversion of asset prices. For comparison we include the optimal strategy for a risk-neutral agent who optimises the mean, i.e., $\gamma(u)=1$.
As seen in the figure, when, say, $q_t=-4$ and $S_t=1.25$, the agent buys $\sim 5$ units of assets when $\mfp=1$, while they sell $\sim 1$ unit when they are risk-neutral. This is because, when $\mfp=1$, the agent is protecting themselves from downside risk and hence aims for $q_t\sim 0$, while the risk-neutral agent simply maximises profit and takes a significant short position.

\section{Conclusions and Future Work}
We pose a generic robust risk-aware optimisation problem, develop a policy gradient approach for numerically solving it, and illustrate its tractability on three prototypical examples in financial mathematics. While the approach appears to work well on a collection of different problems, there are several avenues still open for investigation, such as under what conditions on the RDEU, the controlled rv $X^\phi$, and the uncertainty set $\vartheta_\theta$, is the problem well-posed, as well as establishing the convergence for the policy gradient method itself. We believe that the generality of our proposed RL framework opens doors to help solving a host of other problems, including, e.g., robust hedging of derivative contracts and robustifying optimal timing of irreversible investments. One further issue worth illuminating, is that, while the approach applies to dynamic decision making (such as in Examples \ref{ex:benchmark} and \ref{ex:dynamic}), as RDEU is not a dynamically time-consistent risk measure, the optimal strategies \new{obtained here may be viewed as optimal precommitment strategies}. Hence, there is also need for developing an RL approach for robustifying time-consistent dynamic risk measures.

\bibliographystyle{siamplain}
\bibliography{refs}

\small 

\appendix

\section{Proofs}

\begin{proof}[Proof of Proposition \ref{prop:inner-gradient-formula}]

First, we prove the following lemma.
\begin{lemma}[Representation of RDEU]\label{app:lem-RDEU}
If the distortion function $g$ is left-differentiable, then the RDEU admits representation
\begin{equation}
\RM^U_g[Y] = - \int_0^1 U\left(\Finv{Y}(s)\right) \,\gamma(s) \, ds\,,
\end{equation}
where $\gamma \colon (0,1) \to \R_\ge$ is given by $\gamma(u):= \partial_{-} g(x)|_{x = 1 - u}$, and $\partial_{-}$ denotes the derivative from the left.
\end{lemma}
\begin{proof}[Proof of Lemma \ref{app:lem-RDEU}]
Using first integration by parts, then the properties of $g$, we obtain
\begin{equation}
\begin{split}
    \RM^U_g[Y] 
    &= 
    \Big(1 - g\big(\P(U(Y) > y) \big) \Big) \,y\,\Big|_{-\infty}^0
    + \int_{- \infty}^0y\, d\, \left[g\big(\P(U(Y) > y)\big)\right]
    \\
    & \quad - g\big(\P(U(Y)>y)\big)\, y \,\Big|^{+\infty}_0
    + \int^{+ \infty}_0y \,d\, \left[g\big(\P(U(Y) > y)\big)\right]
    =
    \int_\R y\, d\left[g\big(\P(U(Y) > y)\big)\right]\,.
\end{split}
\end{equation}
Next, noting that $\P(U(Y) > y) = 1 - F_Y(U^{-1}(y))$ and that by assumption $g$ is left-differentiable, we obtain
\begin{align}
    \RM^U_g[Y] 
    &=    
     - \int_\R y \;\partial_{-} g\left(1 - F_Y\left(U^{-1}(y)\right)\right) \,d  F_Y\left(U^{-1}(y)\right)
     \\
    &=    
     - \int_\R y \;\gamma\left( F_Y\left(U^{-1}(y)\right)\right) d \, F_Y\left(U^{-1}(y)\right) 
     = 
    - \int_0^1 U\left(F_Y^{-1} (s) \right)\gamma (s) \, ds\,,
\end{align}
where we used the change of variable $s :=F_Y\left(U^{-1}(y)\right)$.
\end{proof}

Next, as the cost functional for the $p$-Wasserstein distance in one-dimension is submodular,  we may write \cite{ambrosio2003lecture}[Chapter 1.] 
\begin{equation}
 d_p[X\,,\, Y] 
    = \left(\int_0^1 \left|\Finv{X}(u) - \Finv{Y}(u)\right|^p\;du\right)^{1/p}\,.   
\end{equation}
Using \ref{app:lem-RDEU}, we may represent the gradient of the augmented Lagrangian as
\begin{align}
\begin{split}
    \nabla_\theta L[\theta,\phi] =& -\int_0^1 U'\left(F_\theta^{-1}(s)\right)\, \nabla_\theta F_\theta^{-1}(s)\,\gamma(s)\,ds + p\,\lambda\int_0^1 \iota\left(F_\theta^{-1}(s),G^{-1}(s)\right)\,\nabla_\theta F_\theta^{-1}(s)\,ds
    \\
    &+ p\,\mu \left(\int_0^1 (F_\theta^{-1}(s) - G^{-1}(s))^p ds - \ep^p\right)
    \left(\int_0^1 \iota\left(F_\theta^{-1}(s),G^{-1}(s)\right) \, \nabla_\theta F_\theta^{-1}(s)\,ds \right),
\end{split}
\end{align}
where $\iota(x,y):=|x-y|^{p-1}\sgn(x-y)\,\mathds{1}_{d_p(X^\theta,X^\phi)> \ep}$.
As $F_\theta(F_\theta^{-1}(s))=s$, taking gradients wrt $\theta$ provides us with 
\begin{equation}
    \nabla_\theta F_\theta^{-1}(s) = -\left.\nabla_\theta F_\theta(x)|_{x=F_\theta^{-1}(s)}\right/{f_\theta(F_\theta^{-1}(s))}\,.
\end{equation}
Further, if $u\sim\U(0,1)$, then $F^{-1}_\theta(u)\stackrel{d}{=}X^\theta$. The result follows immediately on substituting these expressions and interpreting the integral over $s$ as expectation over a uniform rv.
\end{proof}

\begin{proof}[Proof of Proposition \ref{prop:outer-gradient-formula}]
The proof follows along the same lines as Proposition \ref{prop:inner-gradient-formula}, and uses the envelope theorem \cite{milgrom2002envelope} to evaluate the gradient wrt $\phi$ as gradient of the Lagrangian evaluated at the saddle point obtained by the inner problem, and is omitted for brevity.
\end{proof}

\end{document}